\documentclass[letterpaper]{article} % DO NOT CHANGE THIS
\usepackage[submission]{aaai23}  % DO NOT CHANGE THIS
\usepackage{times}  % DO NOT CHANGE THIS
\usepackage{helvet}  % DO NOT CHANGE THIS
\usepackage{courier}  % DO NOT CHANGE THIS
\usepackage[hyphens]{url}  % DO NOT CHANGE THIS
\usepackage{graphicx} % DO NOT CHANGE THIS
\usepackage{natbib}  % DO NOT CHANGE THIS AND DO NOT ADD ANY OPTIONS TO IT
\usepackage{caption} % DO NOT CHANGE THIS AND DO NOT ADD ANY OPTIONS TO IT
\usepackage{algorithm}
\usepackage{algorithmic}

\usepackage{newfloat}
\usepackage{listings}
\DeclareCaptionStyle{ruled}{labelfont=normalfont,labelsep=colon,strut=off} % DO NOT CHANGE THIS
\floatstyle{ruled}
\newfloat{listing}{tb}{lst}{}
\floatname{listing}{Listing}
\title{Instruct-Video2Avatar: Video-to-Avatar Generation with Instructions}
\author {
    Shaoxu Li,\textsuperscript{\rm 1}
}
\affiliations {
    \textsuperscript{\rm 1} John Hopcroft Center for Computer Science\\
    Shanghai Jiao Tong University\\
    Shanghai, China\\
    lishaoxu@sjtu.edu.cn
}
\usepackage{bibentry}
\begin{document}

%\maketitle

\twocolumn[{%
\renewcommand\twocolumn[1][]{#1}%
\maketitle
\begin{center}
    \centering
    \captionsetup{type=figure}
    \includegraphics[width=0.85\textwidth]{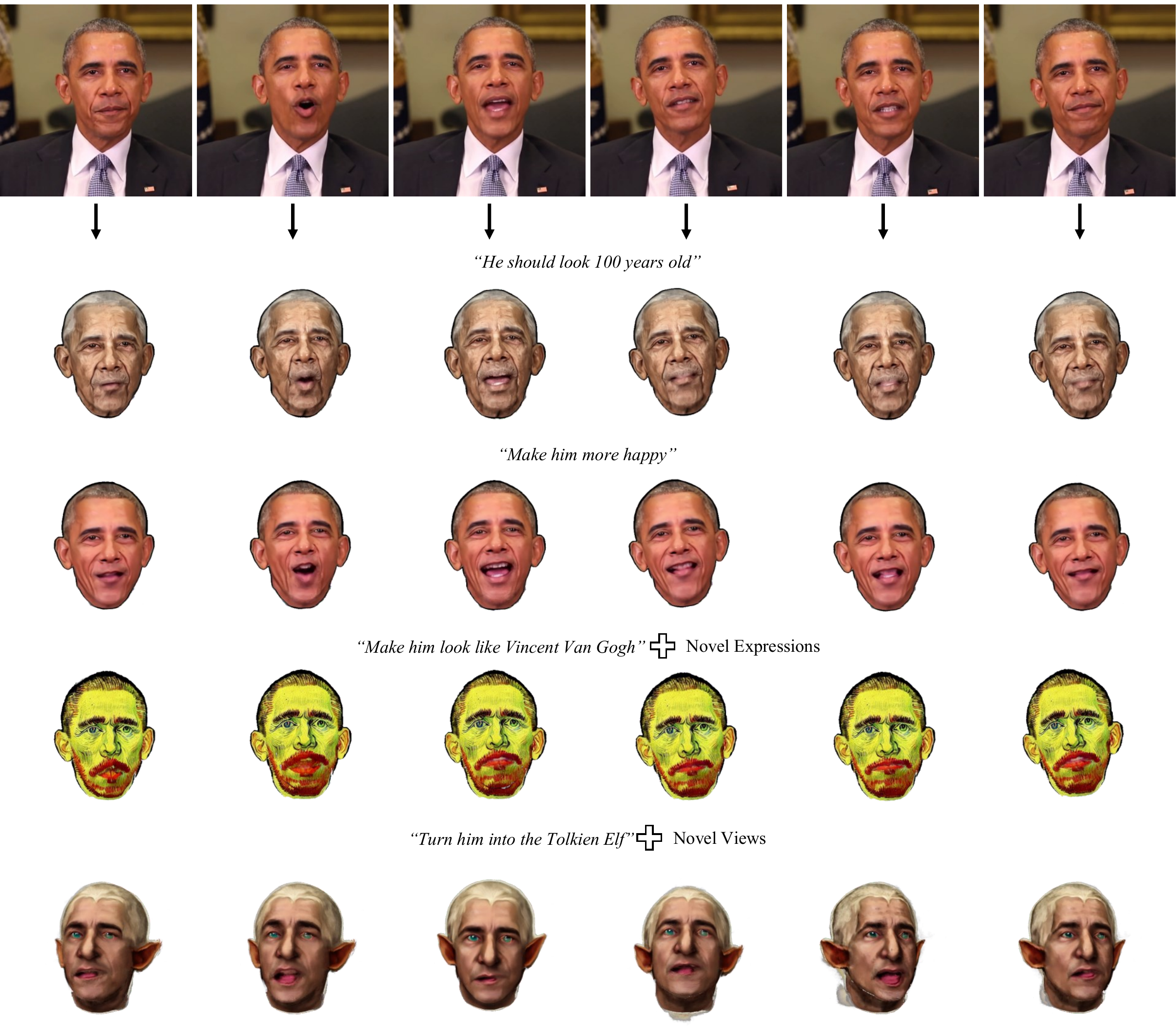}
    \captionof{figure}{Given a short monocular RGB video and text instructions, our method optimizes a deformable neural radiance field to synthesize edited photo-realistic animatable 3D neural head avatars. The resulting head avatar can be viewed under novel views and animated with novel expressions.}
    \label{title}
\end{center}%
}]

\begin{abstract}
We propose a method for synthesizing edited photo-realistic digital avatars with text instructions. Given a short monocular RGB video and text instructions, our method uses an image-conditioned diffusion model to edit one head image and uses the video stylization method to accomplish the editing of other head images. Through iterative training and update (three times or more), our method synthesizes edited photo-realistic animatable 3D neural head avatars with a deformable neural radiance field head synthesis method. In quantitative and qualitative studies on various subjects, our method outperforms state-of-the-art methods.

\end{abstract}

\begin{figure*}[t]
\centering
\includegraphics[width=1\textwidth]{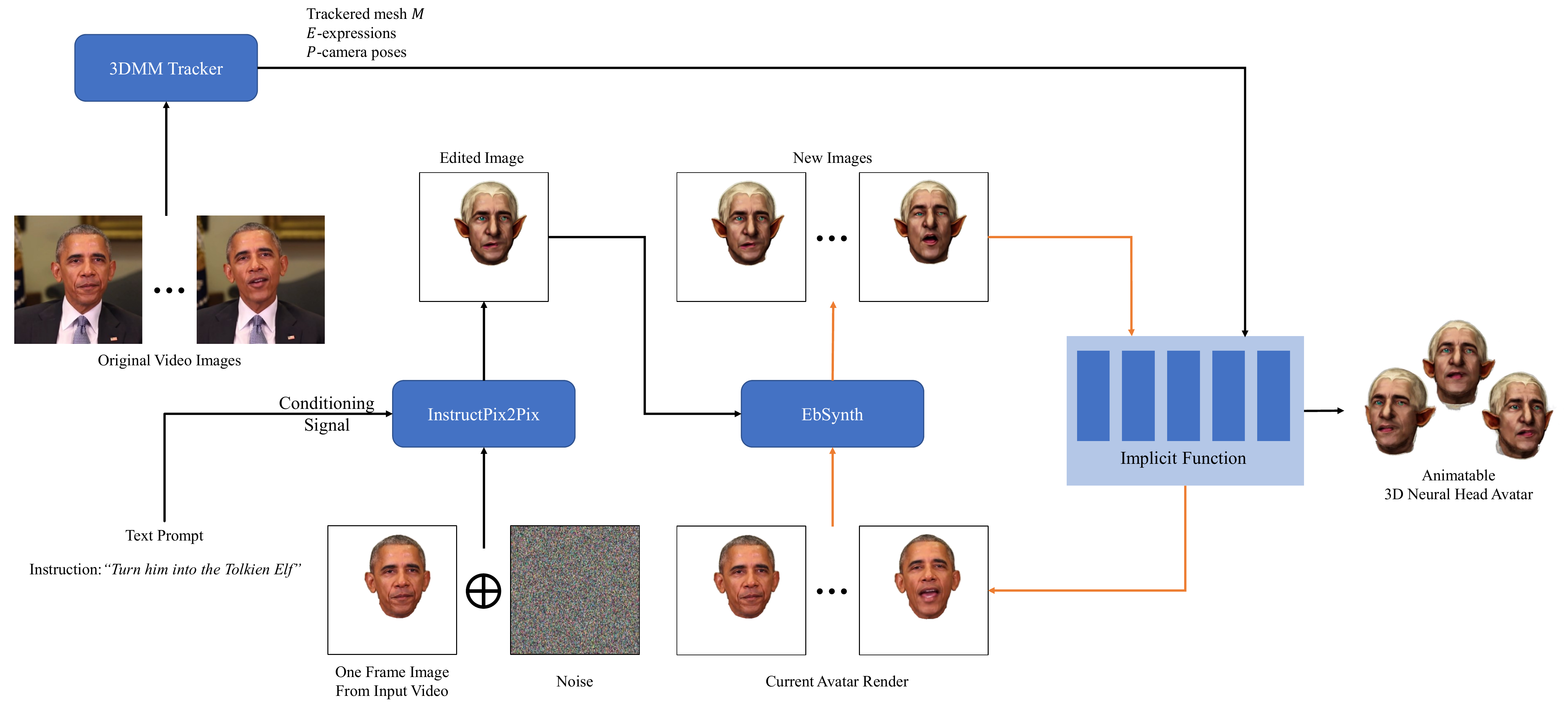} % Reduce the figure size so that it is slightly narrower than the column.
\caption{Overview. Our method gradually updates an animatable 3D neural head avatar by iteratively updating head images: (1) one exampler frame image is selected and edited by InstructPix2Pix given a text instruction, (2) video images are replaced with the edited images by EbSynth, using exampler edited image and original video images, (3) animatable 3D neural head avatar continues training, with all edited images and parameters extracted from the original video images.}
\label{model}
\end{figure*}

\section{Introduction}
With the development of neural 3D head reconstruction techniques, the synthesis and driving of head avatars have been easily appliable. Given a monocular RGB video, typical solutions reconstruct the head poses, camera parameters, and use the head images to optimize a neural radiance field(NeRF)\cite{mildenhall2020nerf}. Although abundant research on photo-realistic animatable 3D neural head avatars\cite{Gafni_2021_CVPRDynamic,zheng2022imavatar,grassal2021neural,ZielonkaCVPR2023INSTA} makes it easy to apply one's photorealistic avatar in VR applications. There is a huge demand on pursuing personality stylization. But the technique that creating one's stylization avatar is still underexplored.

For 3D stylization avatar creation, a typical process needs the modeling artist's operation with professional software. Some user-friendly methods emerged for automatic creation\cite{Han_2021_Exemplar} or interactively creation\cite{shi2019facetoparameter}. Compared with traditional explicit models, implicit 3D representation can achieve higher render quality. But the implicit representation does not have explicit operating nodes like mesh surfaces, which makes manipulation easy. Although some methods try to enable manipulation for NeRFs\cite{Yuan22NeRFEditing,xu2022deforming,li2023interactive}, it's still a tough task to edit NeRFs with user-friendly interaction. The creation of images has been popular with the development of diffusion models\cite{sohldickstein2015deep,rombach2021highresolution,liu2022compositional}. There's an increased demand for creating high-quality 3D scenes easily.

To this end, we propose Instruct-Video2Avatar, a method for synthesizing edited photo-realistic digital avatars that require as input a short monocular RGB video and text instructions. For a head video in the wild, our method can synthesize stylization avatars that integrate the human identity with expected edit instructions, such as "Make him more happy" or "Turn him into the Tolkien Elf", as shown in Figure \ref{title}.

Some text-to-3D methods optimize NeRFs directly with pre-trained 2D diffusion models\cite{kim2022datid3d,wang2023prolificdreamer}. But the quality and stability of scenes are far from commercial use. Some methods optimize existing NeRFs for editing\cite{Haque2023instructnerf,wang2022nerf}, which can achieve expected stylization or transformation. Instruct-NeRF2NeRF\cite{Haque2023instructnerf} is a NeRF-to-NeRF editing method with instruction. Instruct-NeRF2NeRF uses an image-conditioned diffusion model (InstructPix2Pix) to iteratively edit the input images while optimizing the underlying scene. However, we found the editing is time-consuming and inefficient. Inspired by the dataset update of Instruct-NeRF2NeRF, our method gradually updates the radiance field with edited head images. We only use InstructPix2Pix one time and use the video stylization method to edit other images, which is more convenient and faster.

In summary, our contributions are listed as follows:
\begin{itemize}
\item We propose an approach for creating edited photo-realistic animatable 3D neural head avatars, enabling convenient avatar creation with video and text instructions.
\item We propose to update NeRF with video stylization methods, which can be extended to the editing of arbitrarily static or dynamic NeRFs, even for arbitrary video editing.
\item We demonstrate remarkable results of our method through extensive experiments.
\end{itemize}

\section{Related Work}
\paragraph{Image Editing.}
Image editing includes style transfer between artistic styles and translation between image domains. Starting from Gatys et al.\cite{Gatys_2015_Neural}, image style transfer has been deeply studied. Plenty of extended works handled the task of high-quality style transfer\cite{Li_2016_CVPR_Combining,Kolkin_2019_CVPR_Style}, fast style transfer\cite{Johnson_2016_Perceptual,Yao_2019_CVPR_Attention}, geometry aware style transfer\cite{Kim_2020_Deformable}, and applications\cite{Yang_2017_CVPR_Awesome,Azadi_2018_CVPR_Multi}. 

With generative models, many methods manipulate the image by encoding the image into latent space, such as StyleGAN\cite{Karras2019stylegan2}. With pre-trained CLIP\cite{radford2021learning} model, text-to-image models can guide the editing of images with instructions. Recently, diffusion models\cite{sohldickstein2015deep} on image editing have been popular. Latent diffusion models\cite{rombach2021highresolution} offer a convenient way to edit images in latent space. With multiple conditioned diffusion models\cite{liu2022compositional}, methods that create images from multiple inputs emerge. For example, InstructPix2Pix\cite{brooks2022instructpix2pix} can edit an image with an image input and a text instruction.

\paragraph{Video Editing.}
Similar to image stylization, video stylization aims at reproducing the content of a given video using style characteristics extracted from style images. Compared with image stylization, maintaining consistency between different frames is the core task. To deal with this task, some methods propose novel loss functions and networks modules\cite{RuderDB_2016_Artistic,Chen_2017_Coherent,Gao_2020_WACV_Fast} to style a video to the style of an image. Others\cite{Jamriska2018Ebsynth} style the video with an exampler stylized video frame image. For some specific domains, motion retargeting is a hot topic, such as talking head video synthesis. Plentiful works aim to synthesize realistic or artistic talking videos through 2D or 3D methods\cite{hong2022depth,zhang2022sadtalker}. 

Compared with text-to-image, text-to-video is a little tough due to the shortage of high-quality video datasets. Combining pre-trained text-to-image models and attention modules, some methods try to synthesize videos of realistic scenes\cite{wu2021godiva,hong2022cogvideo}. Video diffusion\cite{ho2022video} uses a space-time factorized U-Net with joint image and video data training. Some following works improve the design to generate high-definition videos\cite{Ho2022ImagenVH,wu2022tuneavideo}. But the quality and consistency of synthesis are far from commercial use.

\paragraph{Deformable Neural Radiance Fields.}
NeRF(neural radiance field)\cite{mildenhall2020nerf} can synthesize high-quality novel view images with differentiable volumetric rendering. As an implicit method, the manipulation of the radiance field is not that convenient. A delicate way to edit the radiance field is to warp the NeRF in a canonical space to a deformed space\cite{Yuan22NeRFEditing,xu2022deforming,li2023interactive}. With deformation, NeRF can represent dynamic scenes with temporal and spatial variation. For human facial modeling, a series of approaches have been proposed\cite{Gafni_2021_CVPRDynamic,zheng2022imavatar,grassal2021neural,ZielonkaCVPR2023INSTA}, which leverage 3DMM facial expression code or audio features. NeRFace\cite{Gafni_2021_CVPRDynamic} models the implicit facial model with a low-dimensional 3DMM morphable model which provides explicit control over pose and expressions. In the optimization, the neural radiance field adds the embedded 3DMM expression code to the regular NeRF input. 
IMAvatar\cite{zheng2022imavatar} represents the expression and pose related deformations via learned blendshapes and skinning fields. It employs ray marching and iterative root-finding to locate the canonical surface intersection for each pixel with a novel analytical gradient formulation.
NHA\cite{grassal2021neural} proposes a hybrid representation consisting of a morphable model for the coarse shape and expressions of the face, and two feed-forward networks, predicting vertex offsets of the underlying mesh as well as a view- and expression-dependent texture. And NHA provides a disentangled shape and appearance model of the complete human head (including hair) that is compatible with the standard graphics pipeline.
INSTA\cite{ZielonkaCVPR2023INSTA} models a dynamic neural radiance field based on neural graphics primitives embedded around a parametric face model. INSTA employs a nearest triangle search in deformed space to compute the deformation gradient for mapping between canonical and deformed space.

\paragraph{Editing of Neural Radiance Fields.}
For texture editing of NeRF, many methods emerge leveraging the 2D stylization methods\cite{Chiang2021Stylizing3S,Huang2022StylizedNeRF,Hollein_2022_StyleMesh,li2023instant}. Artistic or realistic texture style transfer can be accomplished with these methods, maintaining the 3D consistency of the NeRF scenes. For geometry-aware NeRF editing, manipulation and generation are the two main tasks. 
Manipulation mainly focuses on the interactive editing of a NeRF scene, leveraging the property of deformable neural radiance fields. These methods operate the scene leveraging on extracting explicit meshes\cite{Yuan22NeRFEditing,xu2022deforming,peng2021CageNeRF} or manipulate the space by some space proxy\cite{peng2021CageNeRF}. The scene generation focuses on creating new scenes with or without an exampler. EditNeRF\cite{liu2021editing} edits a NeRF scene by encoding the scene into latent space.
ClipNeRF\cite{wang2021clip} introduces a disentangled conditional NeRF architecture that allows individual control over both shape and appearance. By leveraging the recent Contrastive Language-Image Pre-Training (CLIP) model, ClipNeRF allows users to manipulate NeRF using either a short text prompt or an exemplar image. 
RODIN\cite{wang2022rodin} proposes the roll-out diffusion network, which represents a neural radiance field as multiple 2D feature maps and rolls out these maps into a single 2D feature plane within which they perform 3D-aware diffusion.
DATID-3D\cite{kim2022datid3d} proposes a novel pipeline of text-guided domain adaptation tailored for 3D generative models using text-to-image diffusion models that can synthesize diverse images per text prompt without collecting additional images and camera information for the target domain.
NeRF-Art\cite{wang2022nerf} introduces a novel global-local contrastive learning strategy, combined with the directional constraint to simultaneously control both the trajectory and the strength of the target style. Moreover, they adopt a weight regularization method to effectively suppress the cloudy artifacts and the geometry noises when transforming the density field for geometry stylization.
Instruct-NeRF2NeRF\cite{Haque2023instructnerf} gradually updates a reconstructed NeRF scene by iteratively updating the dataset images while training the NeRF. Given a NeRF of a scene, Instruct-NeRF2NeRF uses an image-conditioned diffusion model (InstructPix2Pix\cite{brooks2022instructpix2pix}) to iteratively edit the input images while optimizing the underlying scene. Inspired by Instruct-NeRF2NeRF, we propose Instruct-Video2Avatar, which also updates the train images. However, instead of editing all images with a diffusion model, we only edit one exampler image with InstructPix2Pix and edit other images with the video stylization method, which is faster with higher quality.

\begin{figure*}[t]
\centering
\includegraphics[width=1\textwidth]{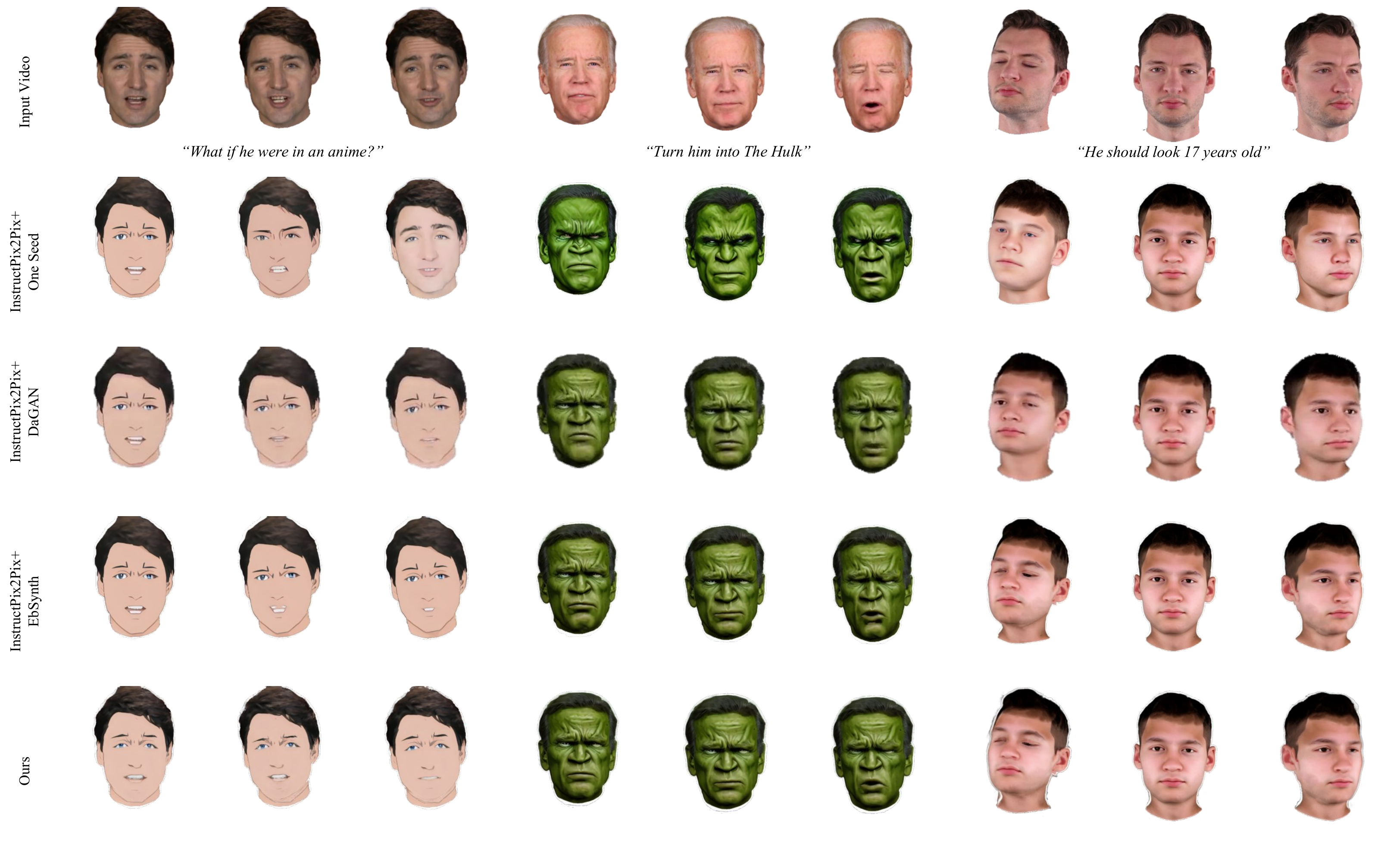} % Reduce the figure size so that it is slightly narrower than the column.
\caption{Comparison with state-of-the-art methods. We compare with One Seed for InstructPix2Pix, DaGAN and EbSynth. The latter two methods use InstructPix2Pix as the basis for clear comparison.}
\label{comparison}
\end{figure*}

\section{Method}
Our method takes as input a talking head video. Additionally, our method takes as input a natural-language editing instruction for avatar generation. As output, our method produces an edited version of a photo-realistic animatable 3D neural head avatar.

Our method segments the head first and iteratively updates the images to accomplish the task. As shown in Figure\ref{model}, our method edits one exampler frame image with a natural-language editing instruction, e.g., "Turn him into the Tolkien Elf". For unedited head images, the editing is accomplished by a video stylization method with the edited image. With all edited images and parameters extracted from the original video images, our method produces a photo-realistic edited animatable 3D neural head avatar. 

Our method builds off three parts, diffusion models for image editing, video stylization from one exampler frame, photo-realistic 3D neural head avatar from a video, specifically InstructPix2Pix\cite{brooks2022instructpix2pix}, EbSynth\cite{Jamriska2018Ebsynth} and INSTA\cite{ZielonkaCVPR2023INSTA}.

\begin{figure*}[t]
\centering
\includegraphics[width=1\textwidth]{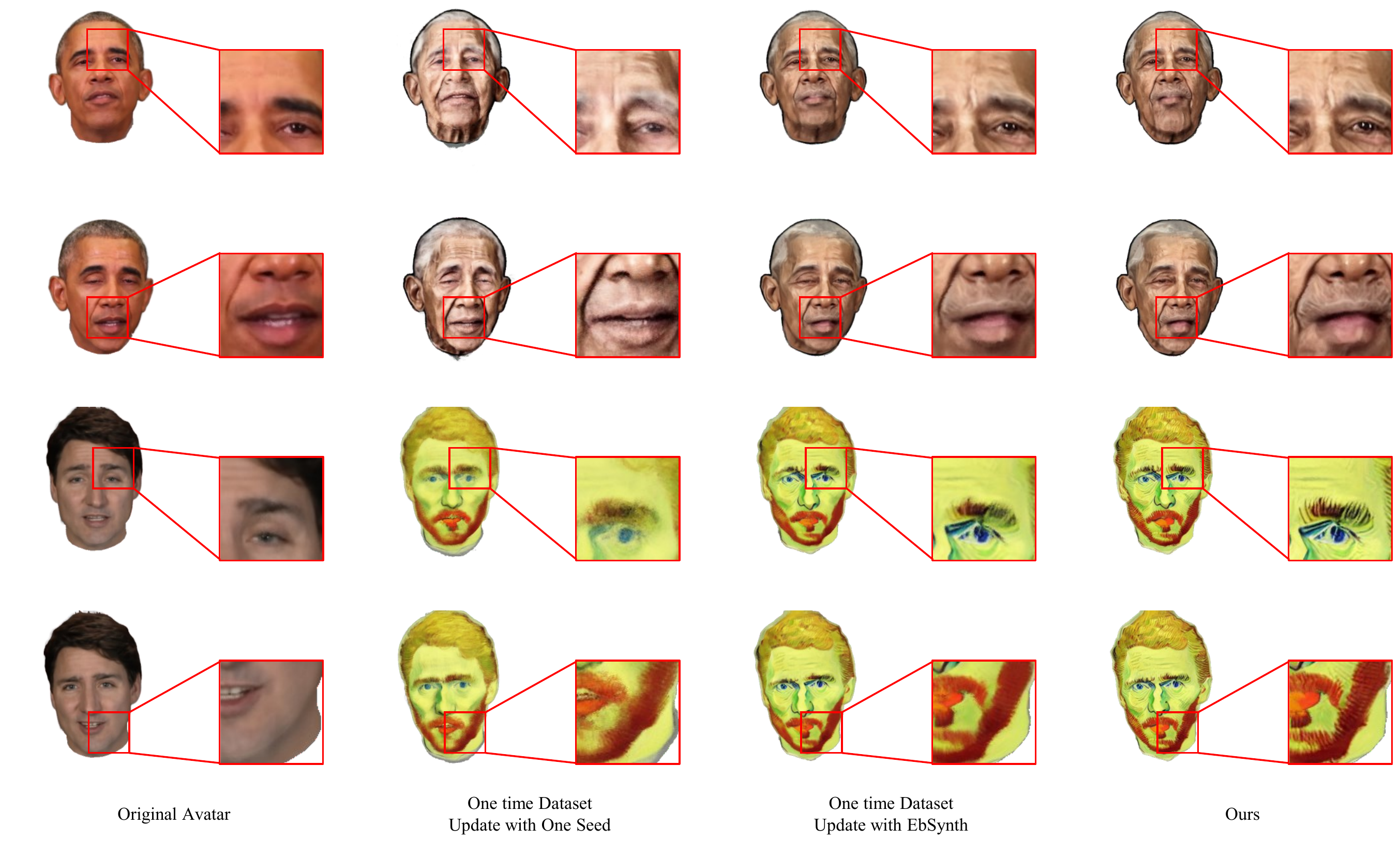} % Reduce the figure size so that it is slightly narrower than the column.
\caption{Baseline Comparisons: We compare our model with variants:One time Dataset Update with One Seed, One time Dataset Update with EbSynth.}
\label{ablation}
\end{figure*}

\subsection{Background}
\subsubsection{Instant Volumetric Head Avatars}
Neural radiance field\cite{mildenhall2020nerf} based head synthesis has attracted much attention for its high quality. INSTA\cite{ZielonkaCVPR2023INSTA} is a neural radiance field based method for photo-realistic digital avatars reconstructing. In practice, INSTA is trained on a single monocular RGB portrait video and can reconstruct a digital avatar that extrapolates to unseen expressions and poses. For a given video, inputs for avatar optimization include head images ${I_i}$, the intrinsic camera parameters $K\in{R^{3\times 3}}$, tracked FLAME\cite{li2017flame} meshes $\{M_i\}$, facial expression coefficient $\{E_i\}$, and head poses $\{P_i\}$. INSTA constructs a neural radiance field in the canonical space to represent the head. The deformation of the radiance field is training for corresponding with facial expressions. Differentiable volumetric rendering is used to optimize the radiance field. With the optimized deformable radiance field, the resulting head avatar can be viewed under novel views and expressions.

\subsubsection{InstructPix2Pix}
Denoising diffusion models have been popular recently. As generative models, an image is gradually formulated from a noisy sample. InstructPix2Pix\cite{brooks2022instructpix2pix} is a diffusion-based method for editing images with text instructions. For editing images with instructions, InstructPix2Pix trains a conditional diffusion model with a paired training dataset of unedited images, edited images, and text instructions. The model is based on Stable Diffusion, the popular text-to-image latent diffusion model. Conditioned on an image $c_I$ and a text editing instruction $c_T$, edited image $z_0$ can be optimized from noise image $z_t$ using the denoising U-Net $\epsilon_\theta$:
\begin{equation}
    \hat{\epsilon} = \epsilon_\theta (z_t;t,c_I,c_T)
\end{equation}
The denoising process predicts the noisy add to the noise image $z_t$, which can be used to estimate edited image $\hat(z_0)$ over timesteps $t\in T$. The variance of edited images increases with timesteps $t$. In practice, InstructPix2Pix is a latent diffusion based method, which operates the diffusion process in latent space. The process of editing an image is encoder-diffusion-decoder. For image editing with two conditions, InstructPix2Pix introduces two guidance scales $s_I$ and $s_T$ to trade off the correspondence between the input image and text instruction. For details, we direct the reader to the original paper\cite{brooks2022instructpix2pix}.

\subsubsection{EbSynth}
Video style transfer aims to implement style transfer with videos, maintaining the stylization quality and the consistency between frames. Exampler-based image video style transfer methods alter the whole video with characteristics from the exampler frame, which has been edited. EbSynth\cite{Jamriska2018Ebsynth} is a synthesizer that can transform full video over an edited frame. EbSynth uses a novel implementation of non-parametric texture synthesis algorithms. EbSynth uses a soft segmentation guide and a positional guide which is essentially a dense warp field that maps every target pixel to its corresponding position in the source. For the task of video stylization, EbSynth can produce stylization images from one frame stylization image. Edited images produced by EbSynth can achieve high quality. The pixel mapping is executed on images, which ignores the 3D geometry. For video stylization, obvious geometric inconsistency can be observed in consecutive video frames.

\subsection{Instruct-Video2Avatar}
Given a short monocular RGB video, neural radiance field based photo-realistic digital avatars can be reconstructed. Our method works by updating the training head images, which maintains consistency between frames. We use a diffusion model to edit one sampler head image, and a video stylization method to edit other images. 

In this section, we first illustrate the process of editing rendered images, then describe the iterative Dataset update strategy, and at last discuss the implementation details of our method. The following mentioned head images are segmented from original videos, which do not include background and body.

\paragraph{Editing Rendered Images}
We use InstructPix2Pix\cite{brooks2022instructpix2pix} to achieve the editing target. As a conditional diffusion model, it takes an unedited image $c_I$, a text instruction $c_T$, and a noisy $z_t$ as input. The text instruction $c_T$ is subjective and the noisy $z_t$ is random. For a head image sequence, we select one sample frame image $I_s$ as the input of InstructPix2Pix. $I_s^e$ denotes the edited image. And then we use EbSynth to edit other images $I_i^e, i\in 1 \cdots N, i\neq s$, with $I_s^e$ as a reference.

One-time image editing with InstructPix2Pix costs about 5 minutes on an RTX 3090 GPU. For editing a NeRF scene, executing editing with diffusion models is time-consuming. For video-to-avatar methods, a video consists of thousands of images, and it is hard to edit images independently like InstructNeRF2NeRF\cite{Haque2023instructnerf}. In addition, the details of the edited images vary with the input images with the same text prompt, even using one seed. NeRF assumes all the images are from one scene. Inconsistent training images make the convergence of the training hard. This leads to more noise and iteration times for training. Considering this, we propose to edit one sampler image $I_s^e$, and use EbSynth to obtain other edited images $I_i^e, i\in 1 \cdots N, i\neq s$.
 
\paragraph{Iterative Dataset Update}
Even though we use EbSynth to ensure consistency between different frames, unnatural distortion from EbSynth lead to noise in the neural radiance field. For high-quality synthesis, we propose an iterative dataset update. We only edit the sampler image once and execute iterations on other images. In the first training, the editing is carried on the head images from the original video. In the later training cycle, the editing is carried out on the rendered images from the optimized head avatar. The neural radiance field promises the 3D consistency of the scene. The EbSynth helps with the rendering quality of the avatar. The whole quality increase with the iteration. After several iterations, we can obtain a photo-realistic edited animatable 3D neural head avatar.

\subsection{Implementation Details}
For InstructPix2Pix, the guidance weights corresponding to the text and image conditioning signals $s_I$ and  $s_T$ can be used to adjust the edit strength. For practicality, the edited frame shall promise the expression the same as the unedited frame. For exampler frame selection, we suggest choosing one head image with mouth open. Editing one frame with mouth open can guide other frames editing better. We suggest using $s_I=1.5$ and  $s_T=3.5$, with 100 denoising steps. The whole process of avatar creation consists of InstructPix2Pix for exampler editing, EbSynth for image editing, INSTA for avatar optimization. It costs less than 20 minutes for one training, excluding the preprocess of the video for facial parameters.

\section{Results}
We conduct experiments on some real scenes preprocessed by INSTA. The size of each dataset (one video) ranges from 2000-3000 images. We execute EbSynth on Windows with the official executable file. Expect for EbSynth, we execute other experiments on Ubuntu equipd a GPU RTX 3090. First, we evaluate our method through qualitative evaluations. To evaluate the design of our approach, we compare our method with some baselines. We also conduct a perceptual study to illustrate the superiority of our method. For better evaluation, we recommend the reader evaluate the performance through the supplemental videos.

\subsection{Qualitative Evaluation}
Our qualitative results are shown in Figure \ref{title} and Figure \ref{comparison}. For different image2image methods, the edited images vary. Our method mainly focuses on the consistency of rendered images of edited avatars. To eliminate the influence of image2image methods, we use InstructPix2Pix as the basis of experiments. For each edit, we compare our method with InstructPix2Pix+One Seed, InstructPix2Pix+DaGAN, and InstructPix2Pix+EbSynth in Figure \ref{comparison}. The first row shows the extracted human head images from the videos, and the latter rows show edited synthesized human head images by different methods.

The edited results of InstructPix2Pix vary with the input image, random seed, and guidance weights. For the results of InstructPix2Pix+One Seed, we fix the random seed and guidance weights. Even though, the edited results are apparent inconsistency due to the poses and expressions of the head. With DaGAN, the edited image consistency increases a lot. But the image quality is inferior and there are significant inconsistencies before and after editing. For example, "The Hulk" can hardly open his mouth and the eyes of the "17 years old man" open unexpectedly. With EbSynth, the edited images are sharpest with good quality and are consistent with the original images. Some noises exist in the edited results. For example, there are noises in the mouth of the "anime man". Our method produces images with good quality. Some shadow noises exist around the avatar head, which are caused by the radiance field. The mouth expressions vary with DaGAN, EbSynth, and our method. It's hard to compare the 3D consistency merely from images. We recommend the reader evaluate the performance through the supplemental videos.

\subsection{Ablation Study}
We validate the architecture of our method by ablation study. The qualitative results are shown in Figure \ref{ablation}. Different from the results in Figure \ref{comparison}, the first column in Figure \ref{ablation} shows the rendered images with training images from the original video. Subsequent columns show corresponding images using different variants. These variants all synthesize 3D neural head avatars, which can be used in virtual reality. Zooming results in red boxes shows the details of image quality, in which our method is superior. The consistency is hard to evaluate from images

\paragraph{One time Dataset Update with One Seed (One time Dataset Update with Modified Instruct-NeRF2NeRF).}
For editing the avatar, we perform one time Dataset Update with one seed, in which all training images are edited by InstructPix2Pix with one seed, and the avatar is trained to fit those edited images. In this way, the optimization process can be regarded as modified Instruct-NeRF2NeRF. Different from the original Instruct-NeRF2NeRF which uses a random noise from a constant range, we fix the noise to decrease the inconsistency. The consistency of edited images determines the quality of the avatar. As a diffusion-based method, the editing results usually have a lot of inconsistency caused by input inconsistency and random seed. For this baseline, we use one seed to enhance the consistency of per-frame editing results. But the inconsistency caused by training image discrepancy exists. Inconsistent training images can lead to blurry and artifact-filled results, as shown in Figure \ref{ablation}. 

\paragraph{One time Dataset Update with EbSynth.}
For editing the avatar, we perform one time Dataset Update with EbSynth, in which one exampler training image is edited by InstructPix2Pix, and the editing is propagated to other training images by EbSynth. Then the avatar is trained to fit those edited images. Compared with one time Dataset Update with one seed, EbSynth significantly enhances the consistency of per-frame editing results. Compared with our pipeline which updates the Dataset three times, the quality of render images is slightly worse, as shown in Figure \ref{ablation}.

\subsection{Perceptual Study}
The task of avatar editing is subjective. We do not have the ground truth. For some editing results, the face shape is far from a normal human. Evaluating the quality of these methods is a challenging task. We recommend the reader evaluate the performance through the supplemental videos. We conducted a perceptual study to evaluate our method. For frame consistency, some methods produce sharp and photo-realistic images which are hard to compare. We present videos generated by our method and other methods to the raters. We ask the raters to rate the given video on "High Definition", and "Temporal Consistency" from 0 (worst) to 5 (best). We recruit 20 participants to score on each question with 10 edited videos. Table \ref{table} shows the results. For "High Definition", InstructPix2Pix+EbSynth gets the highest score and ours is the second-highest. For "Temporal Consistency", ours gets the highest score and One time Dataset Update with EbSynth is the second-highest.

\begin{table}[h]
\centering
\resizebox{1\columnwidth}{!}{
\begin{tabular}{c|c|c}
      & High Definition & Temporal Consistency\\
    InstructPix2Pix+One Seed  &   3.80&   0.95   \\
    InstructPix2Pix+DaGAN&   1.08&   2.41   \\
    InstructPix2Pix+EbSynth&  4.95 &   3.97   \\
    One time Dataset Update with One Seed &   1.05&   2.41   \\
    One time Dataset Update with EbSynth &   2.03&   4.68   \\
    Ours &   4.90&    4.89 
\end{tabular}}
\caption{User study. We present 10 edited videos, each video contains all the methods mentioned above. We ask the participant to score for each method on the "High Definition" and "Temporal Consistency".}
\label{table}
\end{table}

\subsection{Limitations}
\begin{figure}[h]
\centering
\includegraphics[width=0.7\columnwidth]{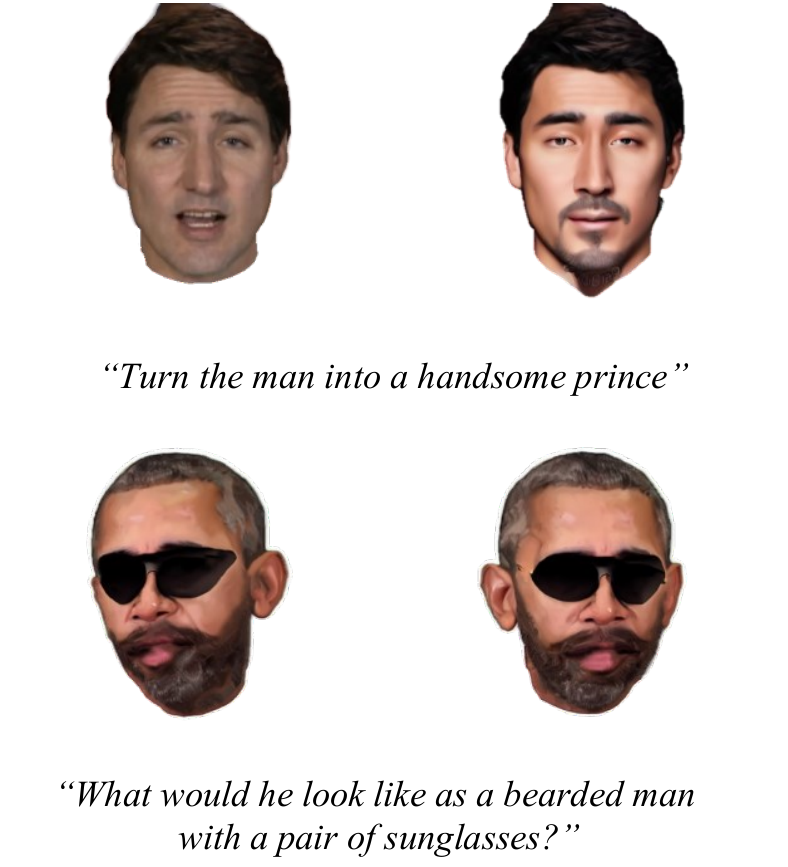} % Reduce the figure size so that it is slightly narrower than the column. Don't use precise values for figure width.This setup will avoid overfull boxes.
\caption{Failure cases: InstructPix2Pix succeeds at editing for the instruction but fails to maintain the expression, and (2) our method succeeds at editing in texture, but the glasses are not independent of the deformable face.}
\label{limitations}
\end{figure}

Our method inherits some limitations of InstructPix2Pix. For some expected editing results, large spatial manipulations may lead to expression inconsistency. INSTA tries to optimize a head avatar for humans, which assumes the surface is close to the face. For some editing instructions that add some objects, our method may produce bad results. We demonstrate two failure cases in Figure \ref{limitations}: (1) InstructPix2Pix succeeds at editing for the instruction but fails to maintain the expression, and (2) our method succeeds at editing in texture, but the glasses is not independent of the deformable face.

\section{Conclusion}
In this paper, we propose Instruct-Video2Avatar, a novel approach that synthesizes edited photo-realistic digital avatars. Our method takes a short monocular RGB video and text instructions as input and uses an image-conditioned diffusion model (InstructPix2Pix) to edit one exampler head image. And then our method uses a video stylization method (EbSynth) to accomplish the editing of other head images. By iterative training and dataset update of the avatar, our method synthesizes edited photo-realistic animatable 3D neural head avatars. In quantitative and qualitative studies on various subjects, our method outperforms state-of-the-art methods. Our method can be extended to the editing of arbitrarily static or dynamic NeRFs, even for arbitrary video editing.

\bibliography{aaai23}

\end{document}